\documentclass[12pt]{article}

\newcommand {\myskip} {\hspace{.15em}}
\newcommand {\fb}[1] {\hbox{\small\bf #1}\myskip}
\newcommand {\Ev} {{\fb E}}
\newcommand {\given} {\; | \myskip}

\renewcommand {\choose}[2] { \left( \!\!\!
                             \begin{array}{c} #1 \\ #2 \end{array}
                             \!\! \right) }

\newcommand {\ave} {\fb {ave}} 
\newcommand{\reals}  {{\rm I\!\! R}}
\newcommand{\calP}   {{\cal P}}
\newcommand{\calN}   {{\cal N}}

\parskip=\medskipamount
\begin{document}

\begin{titlepage}

\begin{center}
{\Large\bf Smoothing Effects of Bagging: ~~~~~~~~~~~~~~~~~~~~~~~~~~~~~~~~~~~~~~ Von Mises Expansions of ~~~~~~~~~~~~~~~~~~~~~~~~~~~~~~~~~~~~~~~Bagged Statistical Functionals} \\~\\~\\
Andreas Buja\footnote{
% AT\&T Labs--Research,
% 180 Park Ave, Florham Park, NJ 07932-0971;
% andreas@research.att.com
  Statistics Department, The Wharton School, University of
  Pennsylvania, Philadelphia, PA, 19104-6340.} \hspace{1cm} Werner
Stuetzle\footnote{Department of Statistics, University of Washington,
  Seattle, WA 98195-4322; wxs@stat.washington.edu. Research partially
  supported by NSF grant DMS - 9803226.} \\
\bigskip \today
\end{center}

\vspace{1cm}

\begin{abstract}
Bagging is a device intended for reducing the prediction error of
learning algorithms.  In its simplest form, bagging draws bootstrap
samples from the training sample, applies the learning algorithm to
each bootstrap sample, and then averages the resulting prediction
rules.  

We extend the definition of bagging from statistics to statistical
functionals and study the von Mises expansion of bagged statistical
functionals.  We show that the expansion is related to the Efron-Stein
ANOVA expansion of the raw (unbagged) functional.  The basic
observation is that a bagged functional is always smooth in the sense
that the von Mises expansion exists and is finite of length 1 +
resample size $M$.  This holds even if the raw functional is rough or
unstable.  The resample size $M$ acts as a smoothing parameter, where
a smaller $M$ means more smoothing.
\end{abstract}

\end{titlepage}

\section{Notations, Definitions and Assumptions for Bagging Statistical Functionals}

We need some standard notations and assumptions in order to define
bagging for statistics and, more generally, for statistical
functionals.

Let $\theta$ be a real-valued statistical functional $\theta(F): \calP
\rightarrow \reals$ defined on a subset $\calP$ of the probability
measures on a given sample space.  By assumption all empirical
measuress $F_M = \frac{1}{M} \sum_{i=1}^M \delta_{x_i}$ are contained
in $\calP$.  If $\theta$ is evaluated at an empiricial measure, it
specializes to a statistic which we write as $\theta(F_M) =
\theta(x_1,\ldots,x_M)$.  This is a permutation symmetric function of
the $M$ sample points.  

In what follows we will repeatedly need expectations of random
variables $\theta(X_1,\ldots,X_M)$ where $X_1,\ldots,X_M$ are
i.i.d.~according to some $F$: 
\[
\Ev_F \, \theta(X_1,\ldots,X_M) = \int
\theta(x_1,\ldots,x_M) \, dF(x_1)\cdots dF(x_M)
\]
Following Breiman~(1996)\nocite{brei96a}, we define bagging of a
statistic $\theta(F_N)$ as the average over bootstrap samples
$X_1^*,\ldots,X_N^*$ drawn i.i.d.~from $F_N$:
\[
\theta^B(F_N) = \Ev_{F_N}\, \theta(X_1^*,\ldots,X_N^*) ~.
\]
For our purposes we need to generalize the notion of bagging to
statistical functionals $\theta(F)$.  First, we need to divorce the
resample size from the sample size~$N$ (compare Friedman and Hall
2000\nocite{frie-hall}, Wu, Goetze, Bickel et al, ... (add more)).  To
this end, we allow the number $M$ of resamples to be drawn from $F_N$
to be arbitrary:
\[
\theta^{B}_M(F_N) = \Ev_{F_N}\, \theta(X_1^*,\ldots,X_M^*) ~.
\]
Note that $M$ is totally independent of $N$; in particular $M$ can be
smaller {\em or} larger than $N$.  This separation of $M$ and $N$
allows us to extend the definition of bagging from empirical measures
$F_N$ to arbitrary distributions:
\[
\theta^B_M(F) = \Ev_F \, \theta(X_1^*,\ldots,X_M^*) ~,
\]
where the random variables $X_1^*,\ldots,X_M^*$ are i.i.d.~$F$, and
their number $M$ is merely a parameter of the bagging procedure.
Unlike for an empirical distribution of an actual sample, for a
general probability measure $F$ there is no notion of sample size.
The variables $X_i^*$ should still be thought of as bootstrap samples,
albeit drawn from an ``infinite population''.

Since the resample size $M$ now denotes a parameter of the bagging
procedure, we need to distinguish it from the size $N$ of actual data
$x_1,\ldots,x_N$.
If one models the data as i.i.d.~samples from $F$, one estimates $F$
with the empirical $F_N = \frac{1}{N} \sum_{i=1}^N \delta_{x_i}$.  The
functional $\theta(F)$ is then estimated by plug-in with the statistic
$\theta(F_N)$:
\[
\hat{\theta}(F) = \theta(F_N) ~.
\]
The bagged functional $\theta_M^B(F)$ in turn is estimated with the
plug-in estimator $\theta_M^B(F_N)$:
\[
\hat{\theta_M^B}(F) = \theta_M^B(F_N) = \Ev_{F_N} \, \theta(X_1^*,\ldots,X_M^*)~.
\]
The idea of bagging is to smooth $\theta$, with the number $M$ playing
the role of a smoothing parameter.  It is not a priori clear, though,
whether more smoothing occurs for small $M$ or large $M$.  Here is an
intuition that proves to be correct: bagging averages over empiricals
$F_M$, hence more smoothing occurs when $F_M$ is allowed to roam
further from $F$, effectively using a larger neighborhood
(``bandwidth'') around $F$; due to $F_M \rightarrow F$ as $M
\rightarrow \infty$, $F_M$ is on the average closer to $F$ for large
$M$, hence the ``bandwidth'' is larger for {\em small} $M$.  The
calculations below verify that this is so, but curiously the reason
has nothing to do with $F_M$ being close to, or far from, $F$: it
turns out that the von Mises expansion of an $M$-bagged functional is
finite of length $M$; because the von Mises expansion is essentially a
Taylor expansion, the $M$-bagged functional is smoother if the
expansion is shorter, that is, if $M$ is smaller.

The above definition of a bagged statistical functional has a blind
spot: It would be interesting to consider both bootstrap sampling with
replacement (conventional) and bootstrap sampling without replacement
where $M$ is strictly smaller than $N$ (as in Friedman and Hall
(2000)\nocite{frie-hall} and Buhlmann and Yu (2000)\nocite{buhl-yu}).
If bootstrap is extended to infinite populations, however, the
difference between sampling with and without replacement disappears.
Thus, in order to capture both modes of sampling, one has to limit oneself to
finite populations and correspondingly to statistics as opposed to
statistical functionals.

If bagging is smoothing by averaging over nearby empirical
distributions, one may wonder whether other types of bagging could be
conceived.  In fact, one can more generally define a smoothed version
$\theta^S$ of $\theta$ by
\[
\theta^S(F) = \ave \{ \, \theta(G) \given G \in \calN(F) \, \} ~,
\]
where $\calN(F)$ is some sort of neighborhood of $F$, and $\ave$
denotes some way of averaging.  This suggests a number of
generalizations of bagging, for example by varying the neighborhoods
and the meaning of $\ave$.  In the present note, however, we remain
with Breiman's original version of bagging and pursue some
implications of averaging over empirical distributions.

\section{Preliminaries 1: The von Mises Expansion of a Statistical Functional}

The von Mises expansion of a functional $\theta$ around a distribution
$F$ is an expansion of the form
\begin{eqnarray*}
\theta(G) &=& \theta(F) + \int \psi_1(x)\,d(G-F)(x) + 
               \frac{1}{2} \int \psi_2(x_1,x_2)\,d(G-F)^{\otimes\,2} + \cdots \\
&=& \theta(F) + \sum_{k=1}^\infty \frac{1}{k!} 
    \int \psi_k(x_1,\ldots,x_k)\,d(G-F)^{\otimes\,k} ~.
\end{eqnarray*}
It can be interpreted as the Taylor expansion of $\theta((1-s)F + sG)$
= $\theta(F + s(G-F))$ evaluated at $s=1$.  The first term in the sum
is a linear functional, the second term is a quadratic functional,
etc.  There is of course no guarantee that the expansion exists.
Reeds~(1976)\nocite{reeds} gives a discussion of conditions under which this
expansion is meaningful in terms of remainders and convergence.  We
are not concerned with technical difficulties because the expansions
we encounter below are finite and exact.  See also Serfling~(1980,
chap.~6)\nocite{serf80}.

The functions $\psi_k$ are not uniquely determined.  We can choose
them such that all the integrals w.r.t.~$F$ vanish, that is,
\begin{eqnarray*}
0 &=& \int \psi_1(x)\,dF \\
0 &=& \int \psi_2(x_1,x_2)\,dF(x_1) = \int \psi_2(x_1,x_2)\,dF(x_2) ~,
\end{eqnarray*}\
and so on.  The von Mises expansion then simplifies to 
\begin{eqnarray*}
\theta(G) 
&=& \theta(F) + \Ev_G \, \psi_1(X) + 
    \frac{1}{2} \, \Ev_G \, \psi_2(X_1,X_2) + \ldots \\
&=& \theta(F) + \sum_{k=1}^\infty \frac{1}{k!} \,
    \Ev_G \, \psi_k(X_1,\ldots,X_k) ~.
\end{eqnarray*}
The function $\psi_1(x)$ is also known as the influence function of
$\theta$, but we will similarly call $\psi_k(x_1,\ldots,x_k)$ the
$k$'th order influence function.  Influence functions of any order are
permutation symmetric in their arguments.

Assuming sufficient smoothness of the functional, $\psi_k$ can be
obtained by differentiation:
\[
\psi_k(x_1,\ldots,x_k) = \left. \frac{d}{ds_1} \right|_{s_1=0} \ldots 
                         \left. \frac{d}{ds_k} \right|_{s_k=0} ~
                         \theta( (1-\sum s_i) F + \sum s_i \delta_{x_i} ) ~.
\]

\section{Preliminaries 2: The ANOVA Expansion of a Statistic}

Efron and Stein (1981)\nocite{es81} introduced an ANOVA-type expansion
for statistics that are functions of independent random variables
$X_1,\ldots,X_M$.  Because we are only interested in symmetric
functions of i.i.d.~data as they arise from evaluating statistical
functionals on empirical distributions, we use an earlier simplified
version of the expansion which can be found for example in
Serfling~(1980)\nocite{serf80}.  Define partial expectations by
\begin{eqnarray*}
\mu_0                  &=& \Ev_F\, \theta(X_1,\ldots,X_M) \\
\mu_1(x_1)             &=& \Ev_F\, \theta(x_1,X_2\ldots,X_M) \\
\mu_2(x_1,x_2)         &=& \Ev_F\, \theta(x_1,x_2,X_3\ldots,X_M) \\
\ldots \\
\mu_k(x_1,\ldots,x_k)  &=& E_F\, \theta(x_1,\ldots,x_k,X_{k+1},\ldots,X_M) \\
\ldots \\
\mu_M(x_1,\ldots,x_M)  &=&  \theta(x_1,\ldots,x_M) ~.
\end{eqnarray*}
Permutation symmetry of $\theta(x_1,\ldots,x_M)$ implies that the free
arguments $x_j$ could be in any position, a fact that will be used
extensively below.  

Define ANOVA terms by
\begin{eqnarray*}
\alpha_0                 &=& \mu_0 \\
\alpha_1(x_1)            &=& \mu_1(x_1) - \mu_0 \\
\alpha_2(x_1,x_2)        &=& \mu_2(x_1,x_2) - \mu_1(x_1) - \mu_1(x_2) + \mu_0 \\
\ldots \\
\alpha_k(x_1,\ldots,x_k) &=& \sum_{\nu=0}^k (-1)^{k-\nu} \sum_{1 \le i_1 < \ldots < i_\nu \le k} 
                           \mu_\nu(x_{i_1},\ldots,x_{i_\nu}) \\
\ldots \\
\alpha_M(x_1,\ldots,x_M) &=& \sum_{\nu=0}^M (-1)^{M-\nu} \sum_{1 \le i_1 < \ldots < i_\nu \le M} 
                           \mu_\nu(x_{i_1},\ldots,x_{i_\nu}) ~.
\end{eqnarray*}
Then the ANOVA expansion of $\theta(x_1,\ldots,x_M)$ is:
\begin{eqnarray*}
\theta(x_1,\ldots,x_M) =&& \alpha_0 + 
                           \sum_{j=1}^M \alpha_1(x_j) + 
                           \sum_{1 \le j_1 < j_2 \le M} \alpha_2(x_{j_1},x_{j_2}) + 
                           \ldots \\
                       =&& \sum_{k=0}^M ~ \sum_{1 \le j_1 < \ldots < j_k \le M} 
                           \alpha_k(x_{j_1},\ldots,x_{j_k}) ~.
\end{eqnarray*}
This expansion is tautological and holds without assumptions other
than permutation symmetry of $\theta(x_1,\ldots,x_M)$ in its
arguments.  The proof is by showing that the partial expectations
implicit in the ANOVA terms cancel each other except for $\mu_M =
\theta(x_1,\ldots,x_M)$.

If one assumes that the variables $X_1,\ldots,X_M$ are i.i.d., then
the terms $\alpha_k$ have vanishing marginals in all arguments:
\[
\Ev_F\, \alpha_k(x_1,\ldots,x_{j-1},X_j,x_{j+1},\ldots,x_k) = 0 ~.
\]
As a consequence, all terms in the ANOVA expansion are pairwise
uncorrelated.

Note that all functions $\mu_k$ and $\alpha_k$ are implicitly
dependent on $M$ because they derive from a statistic of $M$
arguments, $\theta(x_1,\ldots,x_M)$.  If necessary we make the
dependence explicit by writing $\mu_k^M$ and $\alpha_k^M$.  By
contrast, the influence functions $\psi_k$ in the von Mises expansion
are independent of any sample size because this expansion is centered
at $F$ as opposed to $F_M$.

The zero'th term $\alpha_0^M = \mu_0^M$ is also called the ``grand
mean'', and the first term $\alpha_1^M(x)$ the ``main effect
function.''  Correspondingly we call $\alpha_k^M(x_1,\ldots,x_k)$ the
``interaction function'' of $\theta(x_1,\ldots,x_M)$ of order~$k-1$.

\section{A Warm-Up Exercise: The First Order Influence Function of a Bagged Functional}

Before deriving a general formula for the terms of the von Mises
expansion of $\theta^B_M$, we calculate the linear term to illustrate
the idea.  The influence function will be denoted $\psi_1^B(x)$ as a
reminder that it belongs to the bagged functional:
\begin{eqnarray*}
\psi_1^B(x) &=& \left. \frac{d}{ds} \right|_{s=0} \, \theta^B_M((1-s)F + s \delta_x)\\
            &=& \left. \frac{d}{ds} \right|_{s=0} \, \Ev_{(1-s)F + s \delta_x} \, \theta(X_1,\ldots,X_M) ~.
\end{eqnarray*}
The expectation $\Ev_{(1-s)F + s \delta_x} \, \theta(X_1,\ldots,X_M)$ is
effectively a polynomial of degree $M$ in $s$ and hence arbitrarily
differentiable.  We expand it by applying the mixture $(1-s)F + s
\delta_x$ to each argument $X_i$, resulting in $2^M$ terms.  They in
turn can be bundled according to the number of times $\delta_x$
occurs:
\begin{eqnarray*}
&& \Ev_{(1-s) F + s \delta_x} \, \theta(X_1,\ldots,X_M) \\
&& = ~~ (1-s)^M ~ \Ev_F \, \theta(X_1,\ldots,X_M) \\ 
&& ~ + ~ (1-s)^{M-1} \, s ~ M \, \Ev_F \, \theta(x,X_2,\ldots,X_M) \\ 
&& ~ + ~ (1-s)^{M-2} \, s^2 ~ \frac{M(M-1)}{2} \, \Ev_F \, \theta(x,x,X_3,\ldots,X_M) \\ 
&& ~ + ~ O(s^3) ~.
\end{eqnarray*}
Also used was permutation symmetry which implies, for example, 
\[
\Ev_F \, \theta(\ldots,X_{j-1},x,X_{j+1},\ldots) = \Ev_F \, \theta(x,X_2,\ldots,X_M) ~.
\]

As we differentiate w.r.t.~$s$ at $s=0$, only the first two terms make
a contribution:
\[
\psi_1^B(x) = M \, \left[ \, - \Ev_F \, \theta(X_1,\ldots,X_M) 
                             + \Ev_F \, \theta(X_1,\ldots,X_{M-1},x) \, \right] 
            = M \, \alpha_1^M(x) ~,
\]
where as above $\alpha_1^M$ is the main effects function in the ANOVA
expansion of $\theta(F_M)$, which is the raw, not the bagged, statistic.

Suppose we have an i.i.d.~sample $x_1,\ldots,x_N$ of size $N$ from $F$
with empirical distribution $F_N = \frac{1}{N} \sum \delta_{x_i}$.
The first order von Mises approximation to $\hat{\theta^B_M}(F) =
\theta^B_M(F_N)$ is
\[
\theta^B_M(F_N) ~\approx~ \theta^B_M(F) + \frac{1}{N} \, \sum_{i=1}^N \, \psi_1^B(x_i) 
                 ~=~ \mu_0^M + \frac{M}{N} \, \sum_{i=1}^N \, \alpha_1^M(x_i) ~.
\]
For the special case $M = N$ this is exactly the grand mean and the main
effects in the ANOVA expansion of $\theta(F_N)$.

\section{The von Mises Expansion of Bagged Functionals}

We now calculate the $k$-th order influence function.  To this end let
\[
\tilde{F}_k = (1 - \sum_1^k s_i) F + \sum_1^k s_i \delta_{x_i} ~.
\]
By definition,
\[
\psi_k^B(x_1,\ldots,x_k) = 
\left. \frac{\partial^k}{\partial s_1 \cdots \partial s_k} \right|_{s_1,\ldots,s_k = 0} \, 
\theta^B_M(\tilde{F}_k) ~.
\]
Again we note that $\theta^B_M(\tilde{F}_k)$ = $\Ev_{\tilde{F}} \,
\theta(X_1,\ldots,X_M)$ is effectively a polynomial of degree $M$ in
$s$.  Expanding it into $(k+1)^M$ summands, bundling the summands
according to the number of $\delta_{x_i}$'s they contain, and using
permutation symmetry, we get:

\begin{eqnarray*}
&& \theta^B_M(\tilde{F}_k) ~=~ \Ev_{\tilde{F}_k} \, \theta(X_1,\ldots,X_M) \\
&&  = ~ (1-\sum_{i=1}^k s_i)^M \, \Ev_F \, \theta(X_1,\ldots,X_M) \\
&&  ~~~ + \, \sum_{j=1}^k ~ (1-\sum_{i=1}^k \, s_i)^{M-1} \, s_j \, 
                         M \, \Ev_F \, \theta(x_j,X_2,\ldots,X_M) \\
&&  ~~~ + \sum_{1 \le j_1 < j_2 \le k} 
          (1-\sum_{i=1}^k s_i)^{M-2} \, s_{j_1} \, s_{j_2} \, M \, (M-1) \,
          \Ev_F \, \theta(x_{j_1},x_{j_2},X_3,\ldots,X_M) \\
&&  ~~~ + ~ \ldots \\
&&  ~~~ + ~ O(s_1^2,\ldots,s_k^2)
\end{eqnarray*}
Terms containing a second or higher power of any $s_j$ have vanishing
derivatives at zero and hence will disappear in what follows.  This is
why the summation on the fourth line can run over index pairs $j_1 \ne
j_2$ only, the omitted summands being summarily lumped into
$O(s_1^2,\ldots,s_k^2)$.  Thus, with the abbreviated notation for
partial expectations:

\begin{eqnarray*} 
\theta^B_M(\tilde{F}_k) 
&=& \sum_{\nu = 0}^{\min(k,M)} 
    \sum_{1 \le j_1 < \cdots < j_\nu \le k} ~
    (1-\sum_{i=1}^k s_i)^{M-\nu} s_{j_1} \cdots s_{j_\nu} \, \frac{M!}{(M-\nu)!} \,
    \mu_\nu^M(x_{j_1},\ldots,x_{j_\nu}) \\
& & + ~ O(s_1^2,\ldots,s_k^2)~.
\end{eqnarray*}
Note that the outer sum extends to $\min(k,M)$ only.  As the
derivatives can be pulled inside the double sum, we have to calculate
\[
\left. \frac{\partial^k}{\partial s_1 \cdots \partial s_k} \right|_{s_1,\ldots,s_k = 0} 
\left[ (1-\sum_{i=1}^k s_i)^{M-\nu}\, s_{j_1} \cdots s_{j_\nu} \right] ~.
\]
We first take partial derivatives w.r.t.~$s_{j_1},\ldots,s_{j_\nu}$ in turn:
\begin{eqnarray*}
&&  \left. \frac{\partial}{\partial s_{j_1}} \right|_{s_{j_1} = 0}
    \left[ (1-\sum s_i)^{M-\nu} \, s_{j_1} \cdots s_{j_\nu} \right] \\
&&= \left. \left[ (M-\nu) (1-\sum s_i)^{M-\nu-1} (-1) \, s_{j_1} \cdots s_{j_\nu} \,
             + \, (1-\sum s_i)^{M-\nu} \, s_{j_2} \cdots s_{j_\nu} \right] \right|_{s_{j_1}=0} \\
&&= (1-\sum s_i)^{M-\nu} \, s_{j_2} \cdots s_{j_\nu} ~.
\end{eqnarray*}
Repeating this process we obtain:
\[
\left. \frac{\partial^\nu}{\partial s_{j_1} \cdots \partial s_{j_\nu}} \right|_{s_1,\ldots,s_k = 0}
\left[ (1-\sum s_i)^{M-\nu} \, s_{j_1} \cdots s_{j_\nu} \right] ~=~ (1-\sum s_i)^{M-\nu} ~.
\]
We still have to take the derivatives w.r.t.~indices not among
$j_1,\ldots,j_\nu$.  Pick one such index $l$:
\[
\left. \frac{\partial}{\partial s_l} \right|_{s_l = 0} 
\left[ (1-\sum s_i)^{M-\nu} \right] 
= (M-\nu) (1-\sum s_i)^{M-\nu -1} (-1)
\]
Repeating for all such $l$ we get:
\begin{eqnarray*}
&& \left. \frac{\partial^k}{\partial s_1 \cdots \partial s_k} \right|_{s_1,\ldots,s_k = 0} 
   \left[ (1-\sum s_i)^{M-\nu}\, s_{j_1}\cdots s_{j_\nu} \right] \\
&& = \left\{ \begin{array}{l} \displaystyle
 { (M-\nu)(M-\nu-1) \cdots (M-k+1) \, (-1)^{k-\nu} 
   ~=~ \frac{(M-\nu)!}{(M-k)!} \, (-1)^{k-\nu} ~~~{\rm for}~k \le M ~, } \\
 { 0 ~~~{\rm for}~k > M ~. }
	  \end{array} \right.
\end{eqnarray*}
Putting everything together, we get first of all
\[
\psi_k^B(x_1,\ldots,x_k) = 0 ~~~{\rm for}~ k > M ~.
\]
For $k \le M$ we get
\begin{eqnarray*}
\psi_k^B(x_1,\ldots,x_k) 
&=&     \frac{M!}{(M-k)!} \,
        \sum_{\nu = 0}^k (-1)^{k-\nu} 
        \sum_{1 \le j_1 < \cdots < j_\nu \le k}
        \mu_\nu^M(x_{j_1},\ldots,x_{j_\nu}) \\
&=&     \frac{M!}{(M-k)!} ~ \alpha_k^M(x_1,\ldots,x_k)
\end{eqnarray*}
We summarize:

\medskip\noindent {\bf Theorem:} {\em The $k$'th order influence
function $\psi_k^B$ of an $M$-bagged functional $\theta_M^B(F)$ is
proportional to the $k$'th order interaction function $\alpha_k^M$ of
the statistic $\theta(F_M)$:}
\[
\psi_k^B(x_1,\ldots,x_k) ~=~
\left\{ \begin{array}{l} \displaystyle
  { \frac{M!}{(M-k)!} ~ \alpha_k^M(x_1,\ldots,x_k) ~~~{\rm for}~ k \le M ~, } \\
  { 0 ~~~{\rm for}~ k > M ~. }
	\end{array} \right.
\] \medskip

It is now a simple matter to write down the full von Mises expansion
of an $M$-bagged functional:
\begin{eqnarray*}
\theta_M^B(G) 
&=& \theta_M^B(F) \, + \, \sum_{k \ge 1} \, \frac{1}{k!} \, \Ev_G \, \psi_k(X_1,\ldots,X_k) \\
&=& \alpha_0^M    \, + \, \sum_{k=1}^M  \, \choose{M}{k} \, \Ev_G \, \alpha_k^M(X_1,\ldots,X_k) ~.
\end{eqnarray*}
Again we summarize:

\medskip\noindent {\bf Theorem:} {\em Bagged functionals are smooth in
the sense that the von Mises expansion exists and is of finite length
$M+1$:}
\[
\theta_M^B(G) ~=~ \sum_{k=0}^M \, \choose{M}{k} \, \Ev_G \, \alpha_k^M(X_1,\ldots,X_k) ~.
\] \smallskip

Since the von Mises expansion is effectively a Taylor expansion, it is
natural for exact finite expansions to use their length as an inverse
measure of smoothness: the shorter the expansion the smoother the
functional.  With this interpretation and in light of the theorem,
{\em bagging performs more smoothing for smaller $M$.}

Suppose now we have an i.i.d.~sample $y_1,\ldots,y_N$ of size $N$ from
the distribution $F$.  The von Mises expansion of $\theta_M^B$ at $F_N
= \frac{1}{N} \, \sum_1^N \delta_{y_j}$ centered at $F$ is:
\begin{eqnarray*}
\theta_M^B(F_N) 
&=& \sum_{k=0}^M \, \choose{M}{k} \, \frac{1}{N^k} \, 
                    \sum_{1 \le j_1, \ldots, j_k \le N} \, \alpha_k^M(y_{j_1},\ldots,y_{j_k}) ~.
\end{eqnarray*}
Note that the inner sum is unconstrained.  The bagging parameter $M$
is unconstrained w.r.t.~the sample size $N$: $M$ can be chosen to be
smaller or larger than $N$, which raises the question of criteria for
choosing among values for $M$.  This is then just another form of the
problem of smoothing parameter selection.

For the conventional choice $M=N$ one obtains an interesting
comparison with the ANOVA expansion of $\theta(F_N)$:

\medskip\noindent {\bf Theorem:} {\em The terms in the von Mises
expansion of the conventional $N$-bagged statistic $\theta_N^B(F_N)$
form a superset of the terms in the ANOVA expansion of $\theta(F_N)$.}
\begin{eqnarray*}
\theta_N^B(F_N) 
&=& \sum_{k=0}^N \, \choose{N}{k} \, \frac{1}{N^k} \, 
                    \sum_{1 \le j_1, \ldots , j_k \le N} \, \alpha_k^N(y_{j_1},\ldots,y_{j_k}) ~, \\
\theta(F_N) 
&=& \sum_{k=0}^N ~ \sum_{1 \le j_1 < \ldots < j_k \le N} 
                   \alpha_k^N(y_{j_1},\ldots,y_{j_k}) ~.
\end{eqnarray*} 

The inner sums of the first and the second line have $N^k$ and
$({\scriptstyle \stackrel{N}{k}})$ terms, respectively, the difference
being that the first inner sum runs over unconstrained indices, the
second over strictly ordered indices.  The ratio $({\scriptstyle
\stackrel{N}{k}})/N^k$ downweights the inner sum in the first line to
match the smaller number of terms in the second line.

The difference between the raw and the $N$-bagged statistic is that
the latter includes ``diagonal'' terms such as $\alpha_2^N(y_1,y_1)$,
arising from sampling with replacement in the bootstrap procedure.  By
comparison sampling without replacement amounts to a mere permutation
of the data and hence leaves the value of a permutation symmetric
statistic unchanged.

%================================================================

% Convert to bibitem style:


\vspace{-.1in}

\begin{thebibliography}{9}

% \bibitem{brei84a}
% \textsc{L. Breiman and J. H. Friedman and R. Olshen and C. J. Stone} (1984).
% Classification and Regression Trees.
% Belmont, California: Wadsworth.
% }

\bibitem{brei96a} %
\textsc{Breiman, L.} (1996).
Bagging Predictors.
\textit{Machine Learning}
\textbf{26}, 123--140.

\bibitem{buhl-yu} %
\textsc{Buhlmann, P. and Yu, B.} (2002).
Analyzing Bagging.
\textit{Ann. of Statist.}
\textbf{30}, 927--961.

\bibitem{es81} %
\textsc{Efron, B. and Stein, C.} (1981).
The jackknife estimate of variance.
\textit{Ann. of Statist.}
\textbf{9}, 586--596.

\bibitem{frie-hall} %
\textsc{Friedman, J.H. and Hall, P.} (2000).
On Bagging and Nonlinear Estimation.
(Report available from: http://www-stat.stanford.edu/\~{}jhf/\#reports)

% \bibitem{hoeff48}
% \textsc{W. Hoeffding} (1948).
% A Class of Statistics with Asymptotically Normal Distribution.
% \textit{Ann. Math. Statist.}
% \textbf{19}, 293--325.

% \bibitem{chen-hall}
% \textsc{S. X. Chen and P. Hall} (2003).
% Effects of Bagging and Bias Correction on Estimators Defined by Estimating Equations.
% \textit{Statistica Sinica}
% \textbf{xx}, xxx-xxx.

% \bibitem{knight-bass}
% \textsc{K. Knight and G. W. Bassett, Jr.} (2002).
% Second Order Improvements of Sample Quantiles Using Subsamples.
% \textit{xxxxxxx}
% \textbf{xx}, xxx--xxx.

\bibitem{reeds} %
\textsc{Reeds, J. A.} (1976).
On the Definition of von Mises Functionals.
\textit{Ph.D. Dissertation}, Harvard University, Cambridge.

\bibitem{serf80} %
\textsc{Serfling, R. J.} (1980).
Approximation Theorems of Mathematical Statistics.
New York: Wiley.

\end{thebibliography}
\end{document}